\documentclass[conference]{IEEEtran}
\IEEEoverridecommandlockouts
\usepackage[noadjust]{cite}
\usepackage[inline]{enumitem}
\usepackage{amsmath,amssymb,amsfonts}
\usepackage{gensymb}
\usepackage{algorithmic}
\algsetup{linenosize=\tiny}
\usepackage{graphicx}
\usepackage{booktabs}
\usepackage{textcomp}
\usepackage{hyperref}
\usepackage[table]{xcolor}
\usepackage[linesnumbered,ruled]{algorithm2e}
\definecolor{bleudefrance}{rgb}{0.19, 0.55, 0.91}

\ifCLASSOPTIONcompsoc
\usepackage[caption=false,font=normalsize,labelfon
t=sf,textfont=sf]{subfig}
\else
\usepackage[caption=false,font=footnotesize]{subfi
g}
\fi
\newcommand{\norm}[1]{\left\lVert #1 \right\rVert}
\def\BibTeX{{\rm B\kern-.05em{\sc i\kern-.025em b}\kern-.08em
    T\kern-.1667em\lower.7ex\hbox{E}\kern-.125emX}}
  
\begin{document}

\title{Federated Data-Driven Kalman Filtering for State Estimation
\thanks{This work has received [funding | co-funding] from the European Union’s Horizon Europe research and innovation programme [and the Swiss State Secretariat for Education, Research and Innovation (SERI) | the Korea Institute for Advancement of Technology (KIAT)] in the frame of the AutoTRUST project “Autonomous self-adaptive services for TRansformational personalized inclUsivenesS and resilience in mobility” under the Grant Agreement No 101148123.
}
}

\author{
\IEEEauthorblockN{Nikos Piperigkos$^{1,2}$, Alexandros Gkillas$^{1,2}$, Christos Anagnostopoulos$^{1,2}$, Aris S. Lalos$^{1}$}
\IEEEauthorblockA{$^1$Industrial Systems Institute, Athena Research Center, Patras Science Park, Greece\\
$^2$AviSense.AI, Patras Science Park, Greece\\
Emails: \{piperigkos, gkillas, anagnostopoulos\}@avisense.ai, lalos@isi.gr
}
}

\maketitle

\begin{abstract}
This paper proposes a novel localization framework based on collaborative training or federated learning paradigm, for highly accurate localization of autonomous vehicles. More specifically, we build on the standard approach of KalmanNet, a recurrent neural network aiming to estimate the underlying system uncertainty of traditional Extended Kalman Filtering, and reformulate it by the \textit{adapt-then-combine} concept to FedKalmanNet. The latter is trained in a distributed manner by a group of vehicles (or \textit{clients}), with local training datasets consisting of vehicular location and velocity measurements, through a global server aggregation operation. The FedKalmanNet is then used by each vehicle to localize itself, by estimating the associated system uncertainty matrices (i.e, Kalman gain). Our aim is to actually demonstrate the benefits of collaborative training for state estimation in autonomous driving, over collaborative decision-making which requires rich V2X communication resources for measurement exchange and sensor fusion under real-time constraints. An extensive experimental and evaluation study conducted in CARLA autonomous driving simulator highlights the superior performance of FedKalmanNet over state-of-the-art collaborative decision-making approaches, in localizing vehicles without the need of real-time V2X communication.
\end{abstract}


\section{Introduction}
\label{intro}

Autonomous vehicles employ a variety of sensors such as cameras, LiDAR, GNSS (global navigation satellite systems), and IMUs (inertial measurement units) to perceive and interpret their environment. These vehicles are expected to be a fundamental component of future Intelligent Transportation Systems \cite{10247090}. Moreover, vehicles enhance their perception capabilities beyond the individual sensor range through Vehicle-to-Vehicle (V2X) communication and 5G, allowing them to share crucial traffic information. Achieving precise 3D location awareness over time is essential for optimizing autonomous driving performance.
A promising approach for enhancing location or situational awareness is to exploit collaboration among vehicles, either during training or the decision-making phase, relying on V2X information exchange \cite{8359446, 9732063}. This approach becomes even more effective when the uncertainty of sensor measurements can be estimated using data-driven or deep learning techniques \cite{9158529}. KalmanNet \cite{9733186}, a recurrent neural network (RNN) designed to estimate the uncertainty for a single agent through the principles of the Extended Kalman filter (EKF), will be used as a key module, exactly due to its interpretability and efficiency in capturing unknown system dynamics. This work will explore the potential of avoiding raw data exchange between vehicles, while still leveraging the information among connected agents. Specifically, we will investigate transitioning from collaboration during the decision-making phase to collaboration during the training phase, and the potential benefits of it. To be more concise, instead of exchanging raw information, the data collected by a group of vehicles can facilitate a continual learning paradigm in order to estimate sensor measurements uncertainty, thereby enhancing the accuracy of localization over time.  

Collaboration during training usually refers to a distributed scenario where clients jointly train models used for localization applications, using their own local models. In this federated learning (FL) scenario, a global server aggregates the local models and sends back to the clients the global model after some communication rounds \cite{zhang2021survey, 10314010}. FedLoc \cite{9250516,9761235} is a very popular generic framework, focusing mainly on indoor localization scenarios of edge devices. However, despite its benefits, it requires extensive trainable parameters and large datasets, even for simple sequences, lacking the explainability of KalmanNet. Indoor localization based on WiFi measurements is also the focus of related FL works \cite{9148111, 9764747}. Collaboration during the decision-making phase, usually refers to cooperative localization (CL) based on traditional optimization techniques. Understanding the statistics of measurement noise is crucial for enhancing location estimation accuracy \cite{8359446}. Centralized methods, such as those using multidimensional scaling \cite{10342697} or quadratically constrained quadratic problems \cite{10159569}, often either assume the noise covariance is known in advance or set it equal to the identity matrix. More practical distributed approaches, which utilize the concept of covariance intersection \cite{9531527, 9662221}, typically assume the true covariance matrices are known, without addressing how they can be estimated in practice. However, in all cases CL requires raw data exchange in order to localize vehicles, thus resulting in high communication costs and privacy issues.
Thus, the challenge addressed in this paper is to design an explainable data-driven localization architecture that utilize the collaborative nature of FL in order to enhance autonomous vehicles localization.

To address that challenge, this study combines FL with the inherent interpretable KalmanNet architecture. This novel integration promises high performance, due to diverse data shared by cooperating agents, as well as low computational complexity due to the explainable nature of KalmanNet. Moreover, motivated by distributed parameter estimation approaches \cite{sayed}, our method employs the \textit{adapt-then-combine} (ATC) strategy. During the \textit{adaptation} step, each vehicle utilizes its private dataset to train a local KalmanNet model. This model estimates the uncertainty of the specific vehicle's measurements, which is then incorporated into the Kalman filter (KF) solution. In the \textit{combination} step, we aim to develop a robust global KalmanNet model that integrates information across various vehicles operating in diverse environments, effectively learning the underlying system's  dynamics. Furthermore, our approach only requires sharing the weights of the local KalmanNet models. By exchanging and fusing the local models during training, we will achieve better performance than CL approaches. 

Therefore, the main contributions of this study can be summarized as follows:
\begin{itemize}
\item Exploiting the ATC strategy, we reformulate standard KalmanNet to its FedKalmanNet counterpart, enabling the formulation of a highly efficient distributed learning framework for data-driven localization, limiting the need for data sharing and ensuring privacy protection.
\item The newly proposed collaborative training paradigm for autonomous vehicle localization is shown to outperform traditional optimization based CL approaches that require exchanging and fusing raw data in real-time V2X conditions.
\item Extensive numerical evaluations carried out in the renowned CARLA simulator \cite{dosovitskiy2017carla} demonstrate the competitive advantages of the proposed federated data-driven localization approach, in terms of absolute pose error.
\end{itemize}
\subsubsection*{Outline} Section \ref{preliminaries} introduces the preliminaries; Section \ref{FedKalmanNet} presents the proposed federated data-driven localization framework; Section \ref{results} is dedicated to experimental setup and results, while Section \ref{conclusion} concludes this work.

\section{Preliminaries}
\label{preliminaries}



\subsection{System model}

Vehicle $i$ at time instant $t$, needs to autonomously navigate using its own sensor capabilities. Its state is characterized by the 3D position, i.e, $\boldsymbol{x_i^{(t)}} = \begin{bmatrix} x_i^{(t)} & y_i^{(t)} & z_i^{(t)} \end{bmatrix}^T \in \mathbb{R}^3$. Utilizing a suite of visual, satellite, and mechanical sensors, the vehicle can gather both self and relative multi-modal observations or measurements concerning its own state 
and that of a nearby vehicle $ j $. More specifically, we can define the state transition and self-positioning models using data from IMU and GNSS sensors. These models, which are assumed to be degraded by Gaussian noise, denoted as $\mathcal{G}(\mu, \boldsymbol{\Sigma})$, can be expressed as follows \cite{8359446}:
\begin{itemize} 
\item State transition model: 
\begin{align}
    \begin{split}
    \label{CV}
    \boldsymbol{x_i^{(t)}} = f(\boldsymbol{x_i^{(t-1)}} ,\boldsymbol{u_i^{(t)}}) + \boldsymbol{e_i^{(t)}},
    \end{split}
\end{align}
\end{itemize}
where $\boldsymbol{e_i^{(t)}} \sim \mathcal{G}(0, \sqrt{\boldsymbol{R_i^{(t)}}})$. 
Function  $f(\cdot)$  employs a constant velocity motion model: $ f = \boldsymbol{A}\boldsymbol{x_i^{(t-1)}} + \boldsymbol{B}\boldsymbol{u_i^{(t)}} $. Here, $ \boldsymbol{A} = \mathbb{I}_3 $ and $ \boldsymbol{B} = \operatorname{diag}(dt, dt, dt) $. Control input vector $\boldsymbol{u_i^{(t)}} = \begin{bmatrix} u_i^{(x,t)} & u_i^{(y,t)} & u_i^{(z,t)} \end{bmatrix}^T \in \mathbb{R}^3 $ consists of 3D velocity as recorded by the IMU sensor.
\begin{itemize}
    \item Self positioning measurement model:
    \begin{align}
    \label{self_positioning}
        \begin{split} \boldsymbol{\hat{z}_{i}^{(t)}} = \boldsymbol{x_i^{(t)}} + \boldsymbol{n_p}, \boldsymbol{n_p} \sim \mathcal{G}(0, \boldsymbol{\Sigma_p})
        \end{split}
    \end{align}
\end{itemize}

Collaborative decision-making approaches \cite{yang, 10325637} based on traditional optimization, exploit the V2X connectivity links among nearby vehicles by fusing self and relative vehicular measurements, in order to localize ego vehicle and its neighbors. Relative measurements or observations include distance, azimuth and inclincation angles with respect to nearby vehicles, extracted by visual sensors like camera or LiDAR. Instead of real-time measurement transmission and fusion between vehicles in challenging environments, the proposed collaborative learning scheme in the context of FedKalmanNet will perform offline local models aggregation using only self measurements. Afterwards, the trained FedKalmanNet will be exploited by each individual vehicle in order to localize itself highly accurate and much more efficient than a collaborative decision-making approach.

\subsection{Data-Driven Kalman Filtering for state estimation} \label{intro-kalman}
Before we proceed with the presentation of our framework, we will revisit the fundamental equations of the EKF, used for state or location estimation in autonomous driving. This review will help illustrate how standard KalmanNet enhances ego vehicle localization through two key features: the representation of non-linear system dynamics and the estimation of covariance matrices for state and measurement noise using an explainable deep learning approach. 
Therefore, the steps for  estimating state $\boldsymbol{\hat{x}_i^{(t)}} \in \mathbb{R}^3$ and   its covariance matrix $\boldsymbol{\hat{S}_i^{(t)}} \in \mathbb{R}^{3 \times 3}$ using the EKF can be described as follows:
\begin{align}
    \begin{split}
    \label{mean_pred}
    {
    \boldsymbol{\overline{x}_{i}^{(t)}} = \boldsymbol{A}\boldsymbol{\hat{x}_i^{(t-1)}} + \boldsymbol{B}\boldsymbol{u_i^{(t)}}}
     \end{split}
    \\
    \begin{split}
    {
    \boldsymbol{\overline{S}_{i}^{(t)}} = \boldsymbol{A_{i}^{(t)}}\boldsymbol{\tilde{S}_{i}^{(t-1)}}\boldsymbol{A_{i}^{(t)T}} + \boldsymbol{R_{i}^{(t)}}}
     \end{split}
    \\
    \begin{split}
    \label{kalman_gain}
    {
    \boldsymbol{K_{i}^{(t)}} = \boldsymbol{\overline{S}_{i}^{(t)}}\boldsymbol{H_{i}^{(t)T}}\left(\boldsymbol{H_{i}^{(t)}}\boldsymbol{\overline{S}_i^{(t)}}\boldsymbol{H^{(t)T}_{i}} + \boldsymbol{Q_{i}^{(t)}}\right)^{-1}}
     \end{split}
    \\
    \begin{split}
    {
    \label{mean_upd}
    \boldsymbol{\hat{x}_{i}^{(t)}} = \boldsymbol{\overline{x}_{i}^{(t)}} + \boldsymbol{K_{i}^{(t)}}\left(\boldsymbol{\hat{z}_{i}^{(t)}} - g\left(\boldsymbol{\overline{x}_{i}^{(t)}}\right)\right)}
     \end{split}
    \\
    \begin{split}
    {
    \label{sigma_upd}
    \boldsymbol{\hat{S}_{i}^{(t)}} = \left(\mathbb{I} - \boldsymbol{K_{i}^{(t)}}\boldsymbol{H_{i}^{(t)}}\right)\boldsymbol{\overline{S}_{i}^{(t)}}},
    \end{split}
\end{align}
where $\boldsymbol{R_i^{(t)}} \in \mathbb{R}^{3 \times 3}$ is the state transition covariance matrix,  $\boldsymbol{Q_i^{(t)}} \in \mathbb{R}^{3 \times 3}$ denotes the measurement covariance matrix and $\boldsymbol{\hat{z}_i^{(t)}}  \in \mathbb{R}^3$ represents the measurement vector  required to estimate the state or location of $i$. Additionally,  $\boldsymbol{H_i^{(t)}}  \in \mathbb{R}^{3 \times 3}$ corresponds to the jacobian matrix of some generic function $g(\cdot)$ with respect to $\boldsymbol{\overline{x}_i^{(t)}}$. In case where $\boldsymbol{\hat{z}_i^{(t)}}$ contains the GNSS position, i.e., direct measurement of $i$'s state, then $g\left(\boldsymbol{\overline{x}_{i}^{(t)}}\right) = \boldsymbol{\overline{x}_{i}^{(t)}}$ and  $\boldsymbol{H_i^{(t)}} = \mathbb{I}_3$. As such, EKF turns to Kalman filter (KF), i.e, its linear counterpart.

The KalmanNet architecture is designed to estimate the uncertainty matrices of KF algorithm. These include the Kalman gain matrix $\boldsymbol{K_i^{(t)}}  \in \mathbb{R}^{3 \times 3}$, the state transition covariance matrix $\boldsymbol{R_i^{(t)}}$, the predicted state covariance matrix $\boldsymbol{\overline{S}_i^{(t)}}  \in \mathbb{R}^{3 \times 3}$, and the  matrix $\boldsymbol{W_i^{(t)}}  \in \mathbb{R}^{3 \times 3}$. The latter is defined as $\boldsymbol{W_i^{(t)}} = \boldsymbol{H_{i}^{(t)}}\boldsymbol{\overline{S}_i^{(t)}}\boldsymbol{H^{(t)T}_{i}} + \boldsymbol{Q_{i}^{(t)}}$.
Using these estimated matrices, KalmanNet computes the updated state estimate $\boldsymbol{\hat{x}_i^{(t)}}$ and its covariance $\boldsymbol{\hat{S}_i^{(t)}}$ following the standard KF  equations. The network takes as input the current and previous measurement vectors $\boldsymbol{\hat{z}_i^{(t)}}$ and $\boldsymbol{\hat{z}_i^{(t-1)}}$, the previous state estimates $\boldsymbol{\hat{x}_i^{(t-1)}}$, $\boldsymbol{\overline{x}_i^{(t-1)}}$, and $\boldsymbol{\overline{x}_i^{(t-2)}}$, the control input vector $\boldsymbol{u_i^{(t)}}$, and the time interval $dt$. Therefore, the equations of the KF can be reformulated as
    \begin{align}
    \begin{split}
    \label{first_eq_knet}
       \boldsymbol{K_{i}^{(t)}}, \boldsymbol{R_{ i}^{(t)}}, \boldsymbol{\overline{S}_{i}^{(t)}}, \boldsymbol{W_{i}^{(t)}} = KalmanNet_{\theta, i}(\cdot)
    \end{split}\\
    \begin{split}
    \label{local_x}
        \boldsymbol{\hat{x}_{i}^{(t)}} = \boldsymbol{\overline{x}_{i}^{(t)}} + \boldsymbol{K_{ i}^{(t)}}\left(\boldsymbol{\hat{z}_{ i}^{(t)}} - \boldsymbol{\overline{x}_{i}^{(t)}}\right)
    \end{split}\\
    \begin{split}
        \label{local_sigma_upd}
        \boldsymbol{\hat{S}_{i}^{(t)}}  = \boldsymbol{\overline{S}_{i}^{(t)}} -  \boldsymbol{\overline{S}_{i}^{(t)}}\boldsymbol{K_{i}^{(t)}}
    \end{split}
    \end{align}
where $KalmanNet_{\theta, i}(\cdot)$ represents the KalmanNet network, which is used to estimate the corresponding uncertainty matrices.
The architecture of KalmanNet consists of three gated recurrent unit (GRU) layers interconnected with fully connected layers. These layers are structured to progressively estimate the required matrices:
The first GRU layer estimates $\boldsymbol{R_i^{(t)}}$.
The second layer uses the output of the first to estimate $\boldsymbol{\overline{S}_i^{(t)}}$.
The third layer utilizes the outputs of the previous layers to estimate $\boldsymbol{W_i^{(t)}}$.
Finally, $\boldsymbol{\overline{S}_i^{(t)}}$ and $\boldsymbol{W_i^{(t)}}$ are used to compute $\boldsymbol{K_i^{(t)}}$. This sequential structure allows the network to capture the dependencies between these matrices in the Kalman filtering process. KalmanNet is also an efficient network, containing almost 20K parameters.

\section{Federated Data-Driven Localization: FedKalmanNet}
\label{FedKalmanNet}

In this Section, the proposed  FedKalmanNet methodology will be presented. Based on the distributed learning theory \cite{sayed}, the proposed FL scheme can be realized by an ATC strategy, motivated by the fact that each vehicle employs a local KF algorithm utilizing the corresponding KalmanNet, and subsequently the local KalmanNet models are fused at the server side in order to derive a more robust and accurate global KalmanNet. Initially, we will
formulate the general approach of FedKalmanNet and then present the proposed adaptation and combination steps of this methodology.

\subsection{Federated KalmanNet} \label{FL-Kalma}

To establish the FL framework, we consider a network of $\mathcal{N}$ vehicles. 
In this distributed learning framework, each vehicle $i$ participating in the proposed collaborative learning process, utilizes its local dataset $\mathcal{D}_i = \{\boldsymbol{{Z}_i^{1:T_i}}, \ \boldsymbol{X_i^{1:T_i}}\}$, containing an input trajectory as measured over time by its own sensors (GNSS, IMU, etc.), as well as the corresponding ground truth (or target) trajectory. More specifically, $T_i$ is the length of training trajectories, input $\boldsymbol{{Z}_i^{1:T_i}} = \begin{bmatrix}\boldsymbol{z^{(1)}_i} & \ldots &  \boldsymbol{z^{(T_i)}_i}\end{bmatrix} \in \mathbb{R}^{6 \times T_i}$ contains the noisy 3D positions and velocities for the corresponding training trajectory, while target $\boldsymbol{X_i^{1:T_i}} = \begin{bmatrix}\boldsymbol{x^{(1)}_i} & \ldots & \boldsymbol{x^{(T_i)}_i}\end{bmatrix}  \in \mathbb{R}^{3 \times T_i}$ contains the corresponding ground truth 3D trajectory. Note that in order to generate input $\boldsymbol{Z_i^{1:T_i}}$, we add white Gaussian noise to ground truth position and velocity of the training dataset following \eqref{CV} and \eqref{self_positioning}. For simplicity, we assume that each local dataset consists of a single pair of input and ground truth trajectories. 

Each vehicle employs a local KF algorithm and trains its corresponding KalmanNet model using its private dataset, following equations \eqref{first_eq_knet}-\eqref{local_sigma_upd}. The fact that each agent employs only its local dataset to train a local model may lead to limitations in the model's ability to generalize across various environmental conditions. The local KalmanNet model  ($KalmanNet_{\theta,i}(\cdot)$) might only capture the uncertainties in sensor measurements specific to the local environment, hence failing to capture the system dynamics  across different scenarios (e.g., weather conditions, trajectories in rural or urban areas).
To address this limitation, we propose the FedKalmanNet framework. This approach enables agents to collaborate under the coordination of a central server. Through this collaboration, vehicles can learn a more robust KalmanNet model that demonstrates enhanced generalization capabilities across diverse environmental conditions. 



\subsection{Federated Data-driven Localization: Adaptation Step} \label{adapt}
During the adaptation step, each vehicle $i$ employs a local KF which utilizes a local KalmanNet $KalmanNet_{\theta, i}(\cdot)$  to estimate the Kalman gain:
    \begin{align}
    \begin{split}
    \label{first_eq_adapt}
       \boldsymbol{K_{i}^{(t)}} = KalmanNet_{\theta, i}(\cdot)
    \end{split}\\
    \begin{split}
    \label{x_adapt}
        \boldsymbol{\hat{x}_{i}^{(t)}} = \boldsymbol{\overline{x}_{i}^{(t)}} + \boldsymbol{K_{ i}^{(t)}}\left(\boldsymbol{\hat{z}_{ i}^{(t)}} - \boldsymbol{\overline{x}_{i}^{(t)}}\right)
    \end{split}
    \end{align}
In the proposed framework, the local KalmanNet can be trained end-to-end using the local dataset. In more detail,  let $\boldsymbol{\theta_i}$ denote the trainable parameters of the local KalmanNet, and $\gamma_i$ be a regularization coefficient. Each agent employs an $\ell_2$-regularized mean-squared error (MSE) loss to optimize its local model, defined as follows:
\begin{align}
\label{eq:loss}
\ell_i \left(\boldsymbol{\theta_i}\right) = \frac{1}{T_i} \sum_{t=1}^{T_i} \norm{\boldsymbol{\hat{x}_i^{(t)}} \left( \boldsymbol{z_i^{(t)}}; \boldsymbol{\theta_i} \right) - \boldsymbol{x_i^{(t)}}} ^2 + \gamma_i \norm{\boldsymbol{\theta_i}}^2
\end{align}
where $\boldsymbol{\hat{x}_i^{(t)}} \left( \boldsymbol{z_i^{(t)}}; \boldsymbol{\theta_i} \right)$ is the output of the local KF parametrized by $\boldsymbol{z_i^{(t)}}$ and $\boldsymbol{\theta_i} $.
The fact that the KalmanNet model is optimized using only the local dataset of each agent may lead to limitations in the model's ability to generalize across various environmental conditions.
Hence, after all participating vehicles $i \in N$ have updated their local KalmanNet using equation (\ref{eq:loss}), the next step is the combination phase.

\subsection{Federated Data-driven Localization: Combination Step} \label{combine}

The goal of this step is to develop a model that captures underlying system dynamics and accurately estimates covariance matrices for state and measurement noise using local data from the vehicles. Given the structure of KFs, agents upload to the central server only $KalmanNet_{\theta_{n}}(\cdot)$, as defined in equation (\ref{first_eq_knet}). The server then aggregates the local KalmanNets using a fusion rule:
\begin{align}
   KalmanNet_{\theta_{g}}(\cdot) = \sum_{i=1}^N a_i KalmanNet_{\theta_{i}}(\cdot)
   \label{eq:combi}
\end{align}
where $KalmanNet_{\theta_{g}}(\cdot)$ denotes the global KalmanNet and $a_i$ are the combination weights.  Afterwards, the server broadcasts the global KalmanNet model back to all clients. Each vehicle then initializes its local  KalmanNet model (equation (\ref{first_eq_knet})) with the received global model. This iterative process is repeated for $M$ communication rounds, enabling the global model to continuously improve, incorporating diverse data as well as ensuring the privacy of local training datasets.

\begin{figure}
\centering
 \includegraphics[scale=0.28]{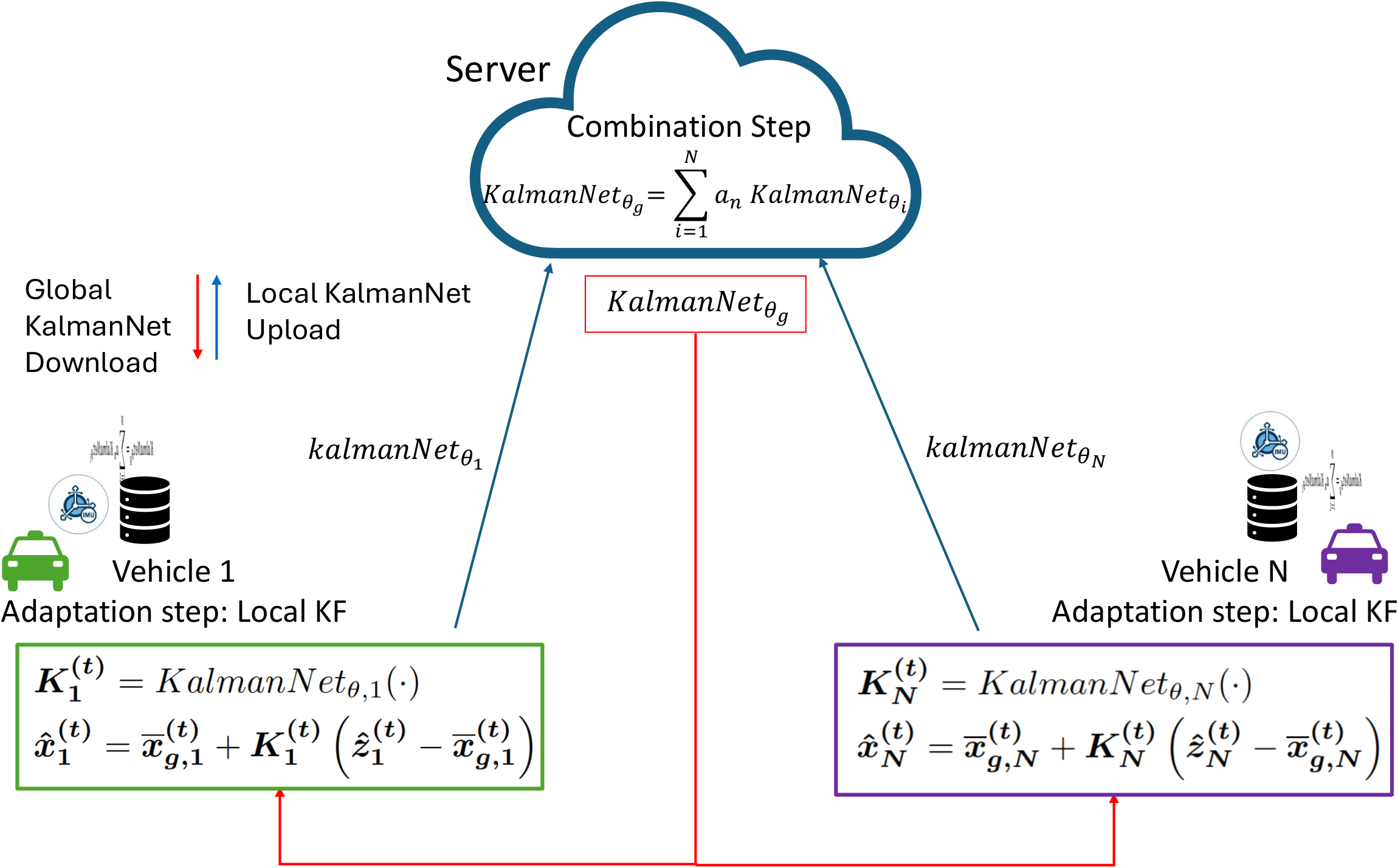}
  \caption{The proposed end-to-end FedKalmanNet approach. The methodology consists of two key steps: adaptation and combination. During the adaptation step, each agent trains the corresponding local KalmanNet model  (\ref{first_eq_knet_FL}) based on KF concept. This ensures that the local model will capture the unique characteristics of each agent's dataset.
During the combination step, the central server aggregates the local KalmanNet models using a fusion rule (\ref{combined_FL}), thereby creating a global model that encapsulates the knowledge from all participating agents.}
  \label{fig:fg_FedKalmanNet}
\end{figure}

Thus, the FedKalmanNet method can be realized by a twofold process: adaptation, i.e., training the local KalmanNet using KF concept and the local private dataset, and combination, i.e., aggregating the local models at the server side and broadcasting the global model back to the agents:
\begin{align}
    \begin{split}
    \label{first_eq_knet_FL}
       \boldsymbol{K_{i}^{(t)}} = KalmanNet_{\theta, i}(\cdot)
    \end{split}\\
    \begin{split}
    \label{local_x_FL}
        \boldsymbol{\hat{x}_{i}^{(t)}} = \boldsymbol{\overline{x}_{i}^{(t)}} + \boldsymbol{K_{ i}^{(t)}}\left(\boldsymbol{\hat{z}_{ i}^{(t)}} - \boldsymbol{\overline{x}_{i}^{(t)}}\right)
    \end{split}\\
    \begin{split}
        \label{combined_FL}
   KalmanNet_{\theta_{g}}(\cdot) = \sum_{i=1}^N a_i KalmanNet_{\theta_{i}}(\cdot)
    \end{split}\\
    \begin{split}
   KalmanNet_{\theta,{i}}(\cdot) = KalmanNet_{\theta_{g}}(\cdot) 
    \end{split}
\end{align}
The proposed FedKalmanNet approach is demonstrated in Figure \ref{fig:fg_FedKalmanNet}. Moreover, Algorithm \ref{algo_fl} summarizes the main steps of the proposed framework.

\begin{algorithm}
\caption{\textbf{FedKalmanNet}}
\begin{algorithmic}
\REQUIRE number of communication rounds $M$, local datasets $\mathcal{D}_i$ for $i=1\dots N$.

\ENSURE \,\, Global KalmanNet

\FOR {each communication round $m=1:M$}
\STATE \textbf{Vehicle side: Adaptation step}

\FOR {each vehicle/client $i=1 \ldots N$}
\STATE Solve the local Kalman Filter

\STATE $\boldsymbol{K_{i}^{(t)}} = KalmanNet_{\theta, i}(\cdot)$
\STATE $\boldsymbol{\hat{x}_{i}^{(t)}} = \boldsymbol{\overline{x}_{i}^{(t)}} + \boldsymbol{K_{ i}^{(t)}}\left(\boldsymbol{\hat{z}_{ i}^{(t)}} - \boldsymbol{\overline{x}_{i}^{(t)}}\right)$

\STATE Optimize the local KalmanNet model using loss function (\ref{eq:loss})
\STATE  Given the structure of Kalman filters, agents upload to the central server only $KalmanNet_{\theta_{i}}(\cdot)$, responsible for Kalman gain estimation. 
\ENDFOR

\STATE \textbf{Server side: Combination step}
\STATE Compute the new  global KalmanNet:\\
$KalmanNet_{\theta_{g}}(\cdot) = \sum_{i=1}^N a_i KalmanNet_{\theta_{i}}(\cdot)$
\STATE Broadcast the global model to all vehicles

\ENDFOR 

\end{algorithmic}
\label{algo_fl}
\end{algorithm}

\section{Numerical results}
\label{results}

\subsection{Simulation setup}
\begin{figure}[htbp]
  \centering
  \subfloat[Client 1]{\includegraphics[width=0.43\linewidth]{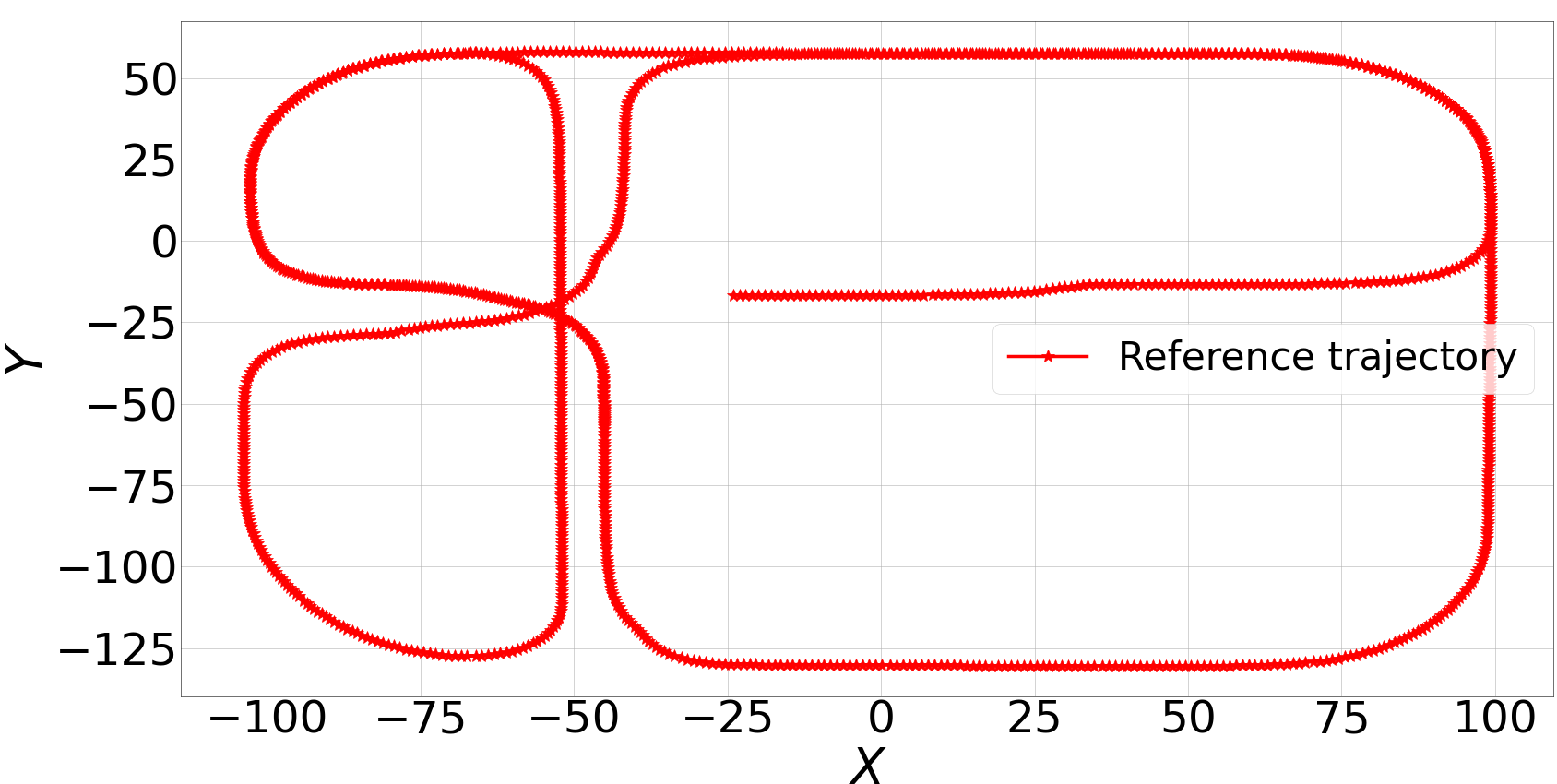}}
  \hspace{2mm}
  \subfloat[Client 2]{\includegraphics[width=0.43\linewidth]{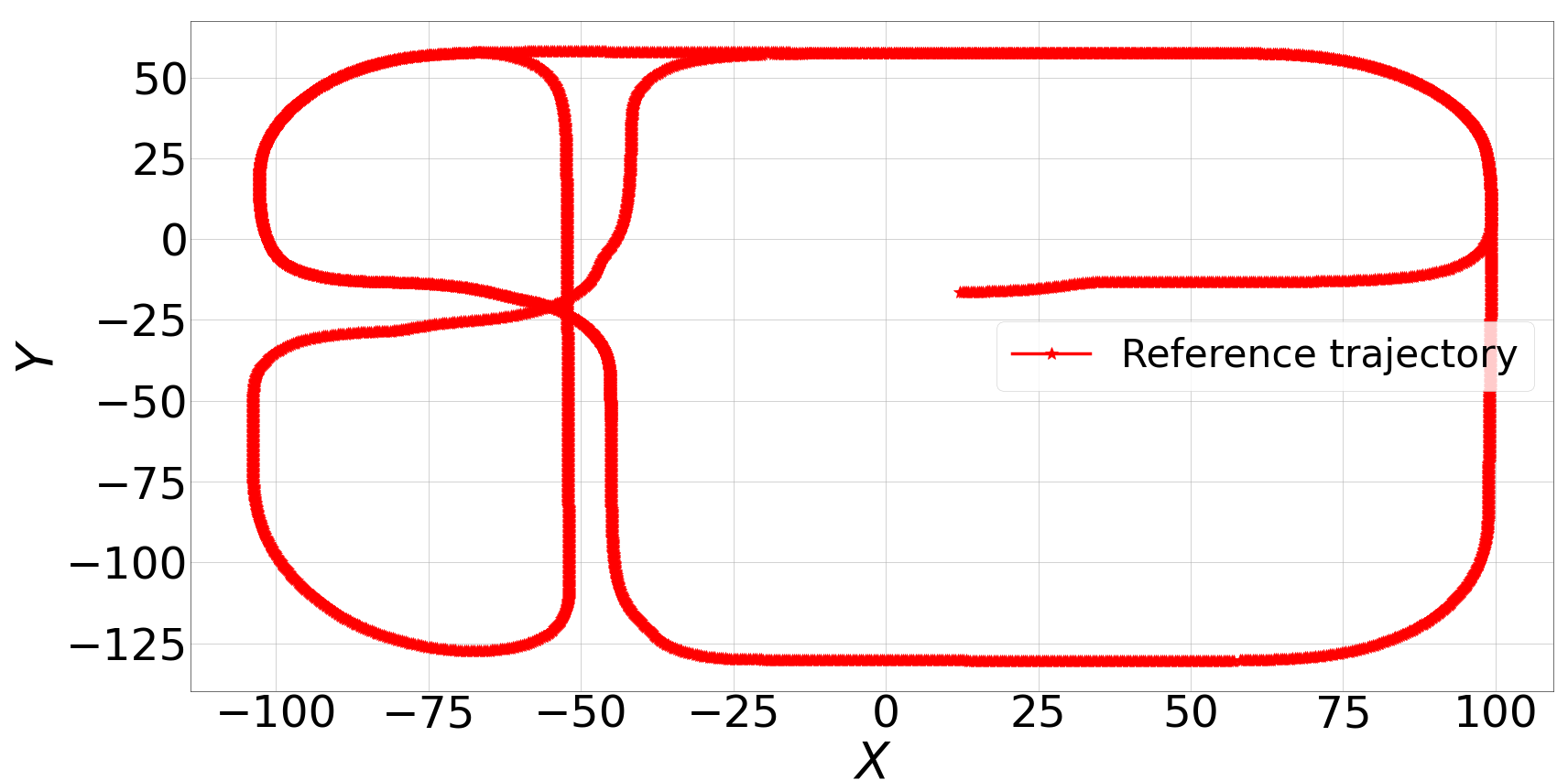}}\\
  \subfloat[Client 3]{\includegraphics[width=0.43\linewidth]{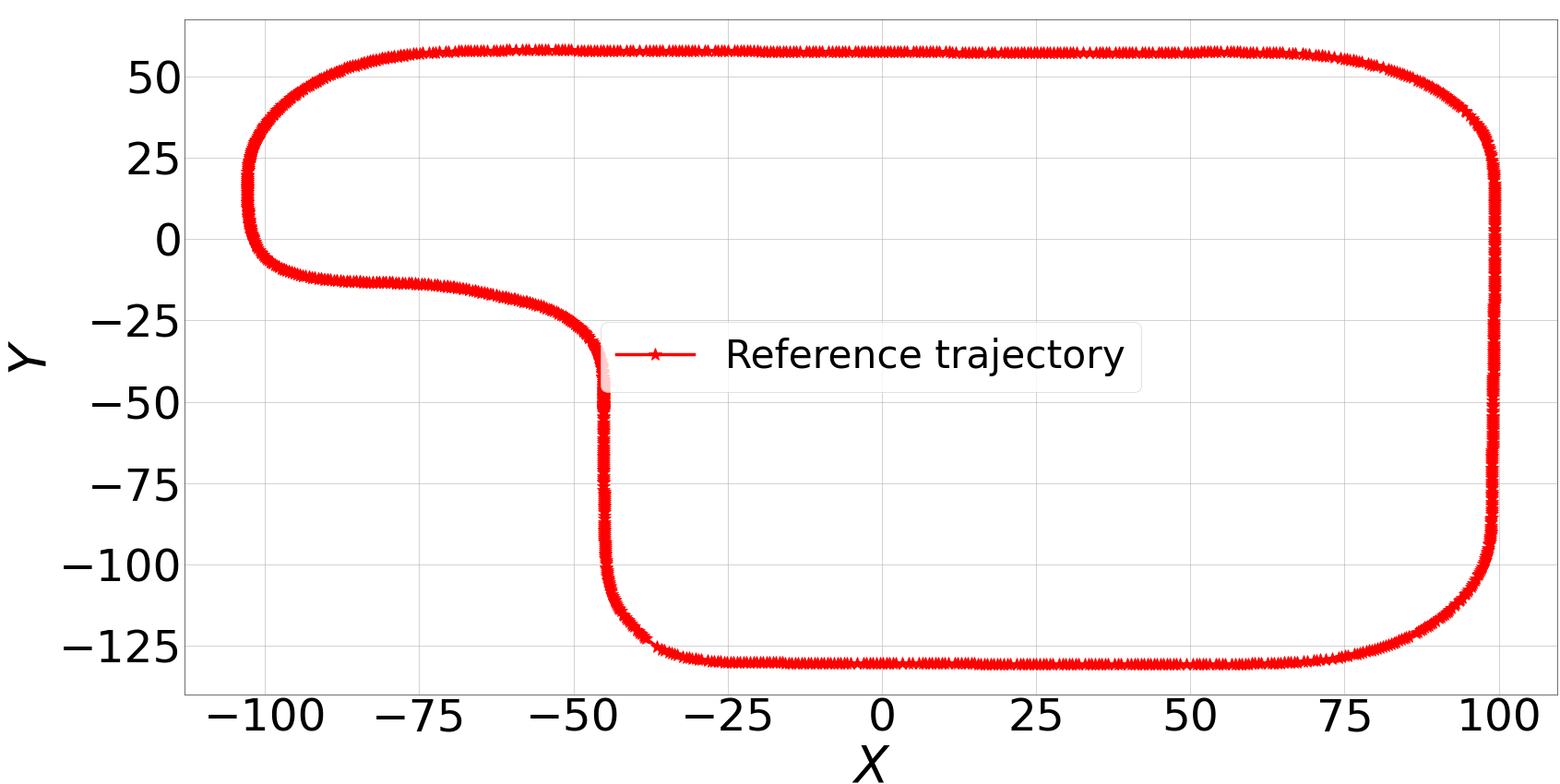}}
  \hspace{2mm}
  \subfloat[Client 4]{\includegraphics[width=0.43\linewidth]{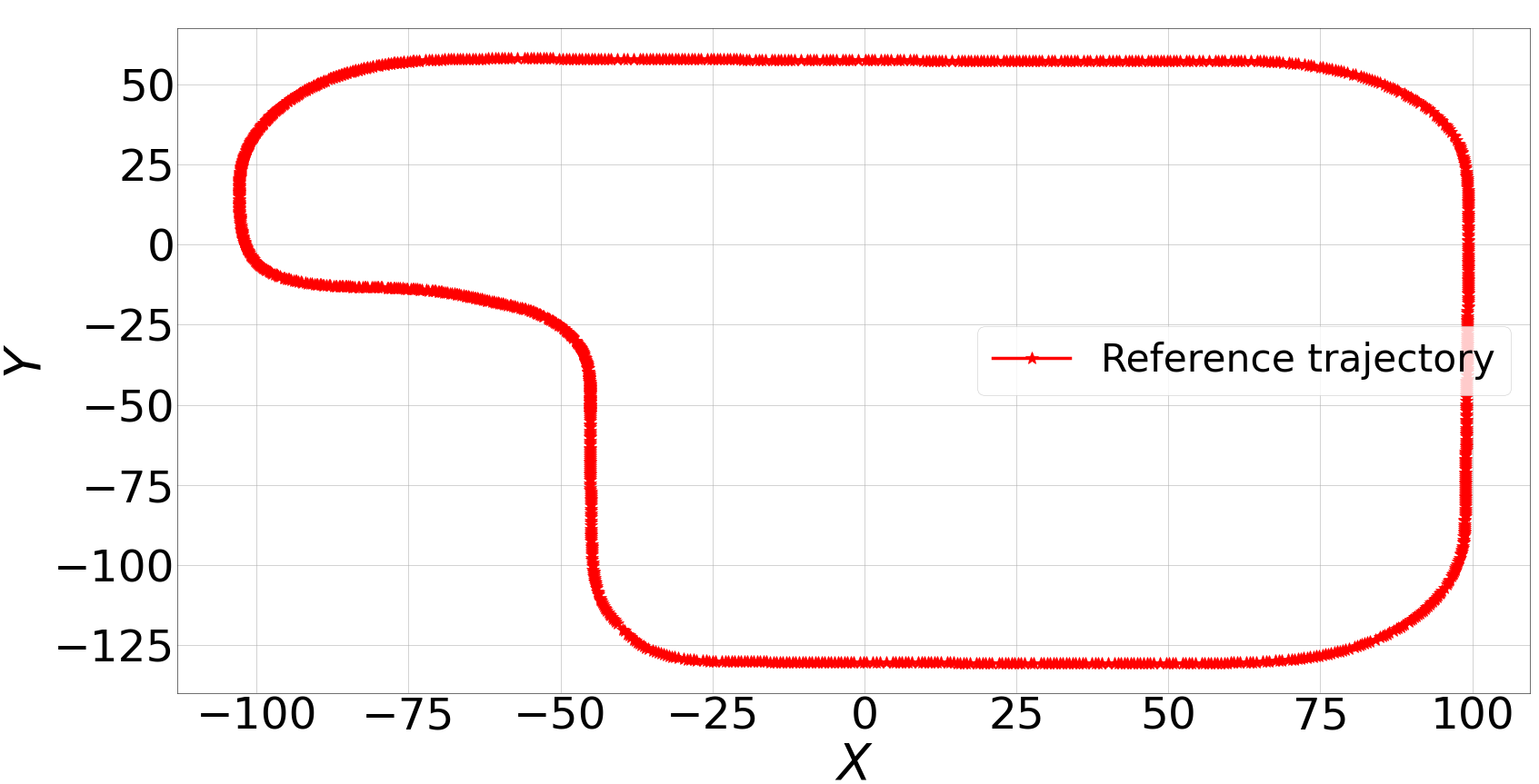}}
   \caption{Clients' trajectories from TownMap10 of CARLA}
   \label{fig:agents}
\end{figure}

The simulations were carried out using dataset \footnote{\url{https://dx.doi.org/10.21227/511y-4s83}} which contains the trajectories of 60 vehicles moving in CARLA simulator's environment \cite{9797115}. The dataset contains ground truth 3D position, linear velocity, acceleration, etc. For the testing evaluation, we have chosen vehicle with index 0 as the ego vehicle 
over the simulation horizon of $T = 900$ time instances and sampling interval $dt  = \ 0.1 \sec$.
The evaluation study will consider two scenarios in order to assess the proposed FL approach: i) comparing \textbf{FedKalmanNet} with the traditional \textbf{KalmanNet} when all the training data are available to the global server, as well as when only the dataset from an individual agent is used during training, ii) how the collaborative training paradigm presented in this work can outperform traditional optimization based CL approaches, which require much larger amount of information from nearby vehicles to localize ego vehicle. Baseline methods, apart from standalone GNSS, include \textbf{MSMV} \cite{yang} and \textbf{LKF-SA} \cite{10325637}, which set the covariance matrix of sensor measurements equal to identity. 
The error metrics include Root Mean Square Localization Error over Time, i.e., $RT-LE$, in order to evaluate ego vehicle $i$'s ability to localize itself. 
Additionally, the cumulative distribution function (CDF) of instantaneous localization errors  demonstrates the probability of location error to be lower or equal than a specific threshold. 
For the training of networks, we have exploited four vehicles/clients with trajectories from TownMap10 of CARLA, and varying lengths ($T = 1550 , \ 4600 , \ 1450 , \ 1420 )$, 
which are shown in Fig. ~\ref{fig:agents}. Although they seem similar, vehicles actually move with different velocities. For example, the minimum and maximum velocity of agents $1$ and $2$ are between $8.9-10.8 \ m/sec$ and  $9.2-10.9 \ m/sec$, respectively, while those of $3$ and $4$ are between $0.7-10.7 \ m/sec$ and  $3.6-10.7 \ m/sec$, respectively. The input to the network is the ground truth 3D trajectory degraded by additive white Gaussian noise of zero mean and standard deviation $\boldsymbol{\Sigma_p} = diag(1.5m, 1.5m, 1.5m)$. Furthermore, additive white Gaussian noise is used for the 3D velocity inputs in order to realistically capture state transition noise. The standard deviation of velocity noise is set to $10\%$ of the ground truth velocity of the specific vehicle, as stated in \cite{Elazab2017}. Each trajectory is splitted in subtrajectories of length equal to $100$, while $80\%$ and $20\%$ out of them are used for training and cross validation. For the centralized training, all four datasets are used to train \textbf{CentrKalmanNet} for 1500 epochs. For the individual training, we exploit the trajectories only from agent $0$ and train the network for 500 epochs. For the FL implementation, we adopt the FedAvg \cite{pmlr-v54-mcmahan17a} implementation, and use 20 communication rounds. Batch size is equal to $1$, while learning rate and weight decay are set to $0.3$.
\subsection{Evaluation study}

\subsubsection{Impact of federated learning vs centralized and individual training}
\begin{figure}
\centering
 \includegraphics[scale=0.16]{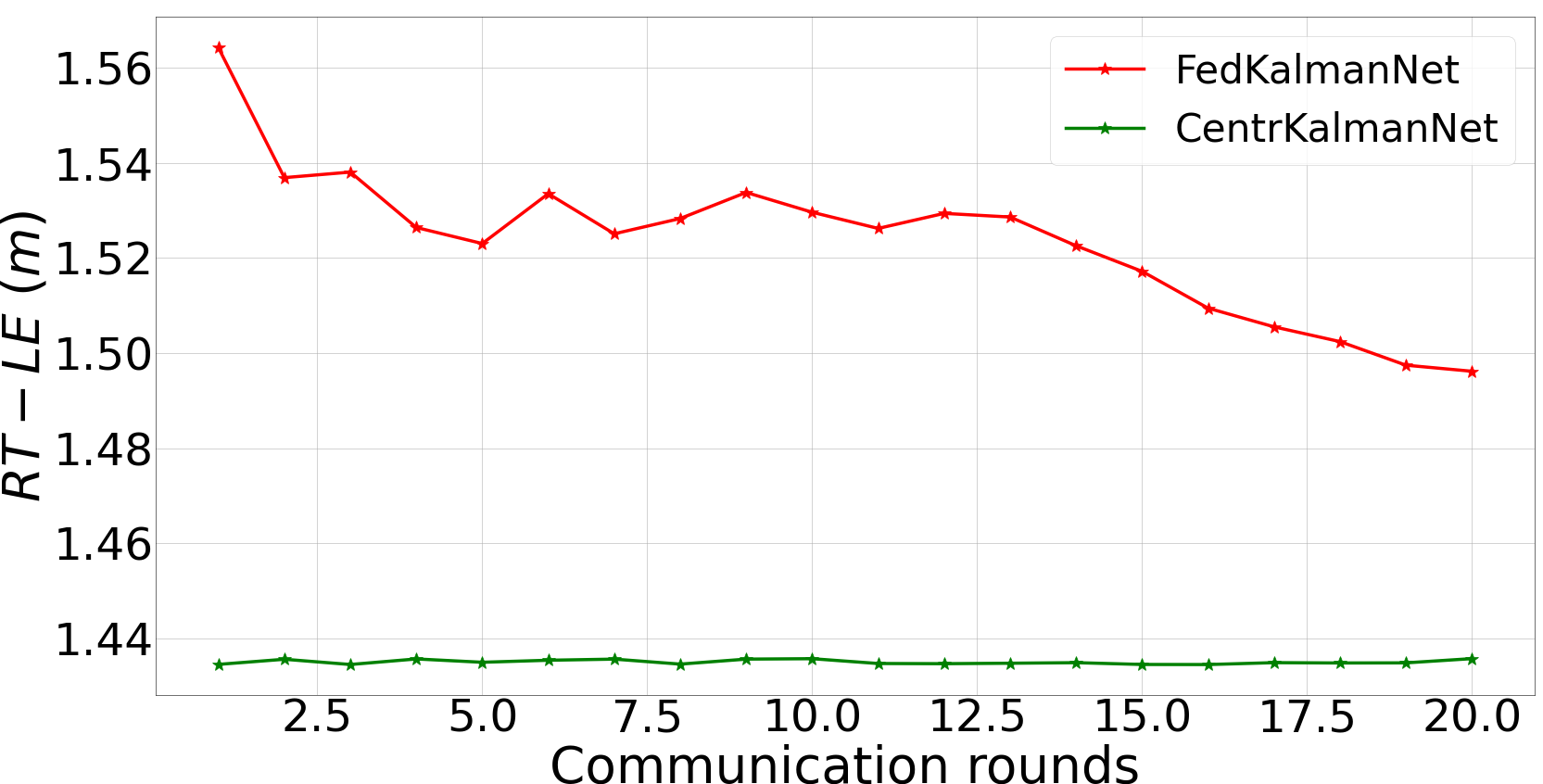}
  \caption{Convergence of \textbf{FedKalmanNet} to \textbf{CentrKalmanNet} after 20 communication rounds  }
  \label{fig:convergence}
\end{figure}

\begin{figure}
\centering
 \includegraphics[scale=0.10]{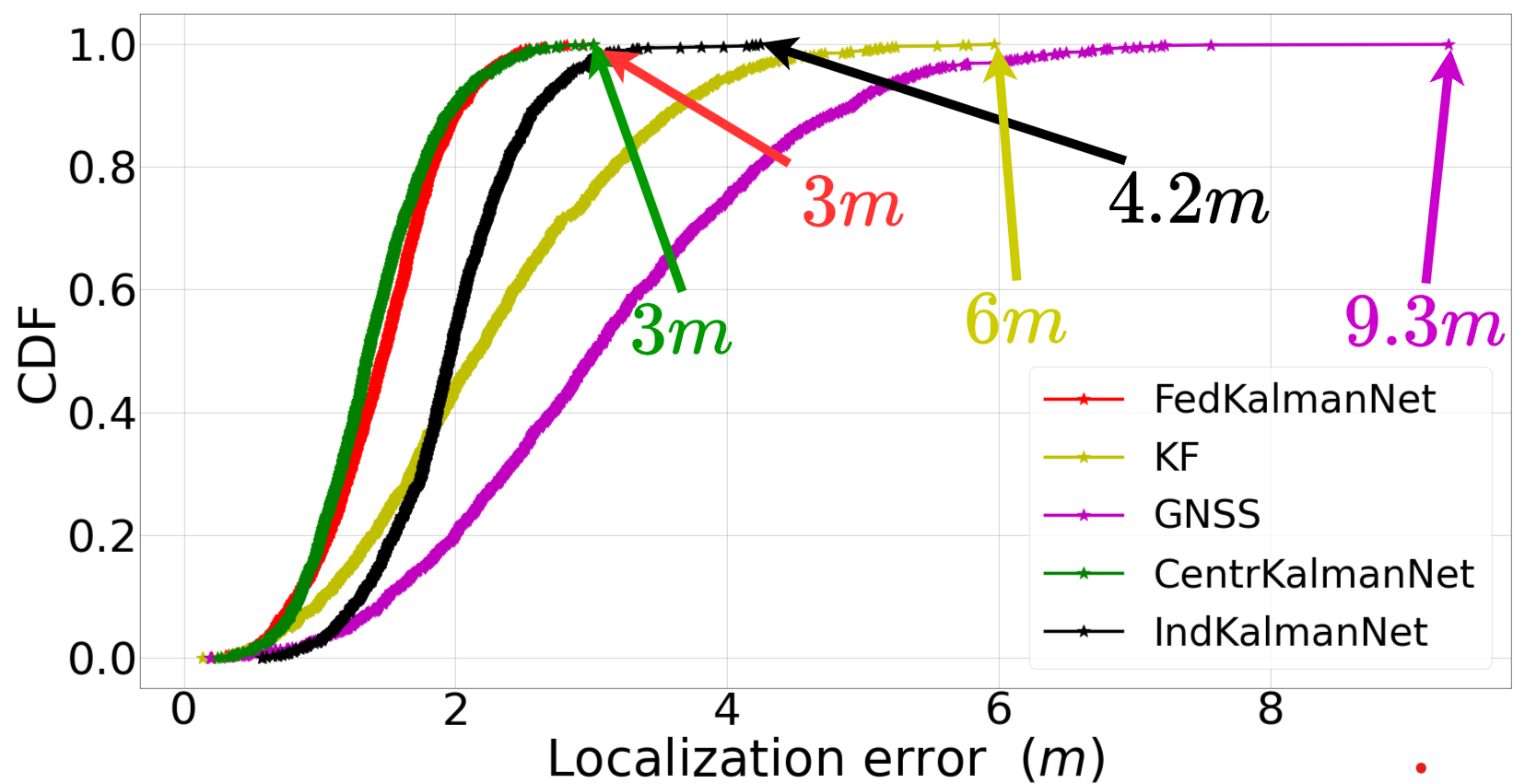}
  \caption{Cumulative distribution function of ego vehicle localization accuracy }
  \label{fig:CDF}
\end{figure}

In this testing scenario, we will evaluate the performance of the proposed data-driven FL framework with respect to centralized and individual implementation of \textbf{KalmanNet's} training. More specifically, Fig.~\ref{fig:convergence} demonstrates the convergence of \textbf{FedKalmanNet} to \textbf{CentrKalmanNet} after 20 communication rounds. To be more detailed, after each round both \textbf{FedKalmanNet} and \textbf{CentrKalmanNet} are evaluated in terms of $RT-LE$ with the GNSS based trajectory of ego vehicle as input, which has been generated by adding white Gaussian noise of zero mean and standard deviation $\boldsymbol{\Sigma_p} = diag(1.8m, 1.8m, 1.8m)$, as well as a bias drawn from uniform distribution $\mathcal{U}[0.5,1]$ to the ground truth. In that way, we will show that the performance of networks is still high enough, regardless of the fact that we have conducted training with input trajectory degraded by Gaussian noise. Furthermore, the standard deviation of velocity noise is set to $15\%$ of the ground truth velocity. Clearly, centralized training achieves superior performance in terms of $RT-LE$ due to the availability of all training data. However, the distributed framework of \textbf{FedKalmanNet} reduces $RT-LE$ after each round, as it is expected from the relevant theory, reaching $RT-LE$ at round 20 lower than $1.5m$, with respect to $1.43m$ of \textbf{CentrKalmanNet}. Additionally, Fig.~\ref{fig:CDF} highlights the CDF of ego vehicle localization error using \textbf{FedKalmanNet} (after 20 rounds of training), \textbf{CentrKalmanNet}, \textbf{IndKalmanNet}, a traditional KF taking as input the GNSS and without any knowledge of system uncertainty, and, finally, GNSS. For each one of the curves, we indicate the maximum error attained by each approach. For example, we see that the simple KF reduces GNSS error by almost $3m$, while the \textbf{IndKalmanNet} reduces it by $5m$, clearly showing the benefits of estimating the underlying uncertainty. Most importantly, the proposed \textbf{FedKalmanNet} reduced maximum GNSS error by $6.3m$, reaching the same accuracy with that of \textbf{CentrKalmanNet}. As such, we conclude that the proposed data-driven FL framework efficiently converges to the centralized model's accuracy, significantly improving at the same time the localization accuracy of ego vehicle.

\subsubsection{Impact of collaborative training vs collaborative decision-making for ego vehicle localization}
\begin{figure}
\centering
 \includegraphics[scale=0.17]{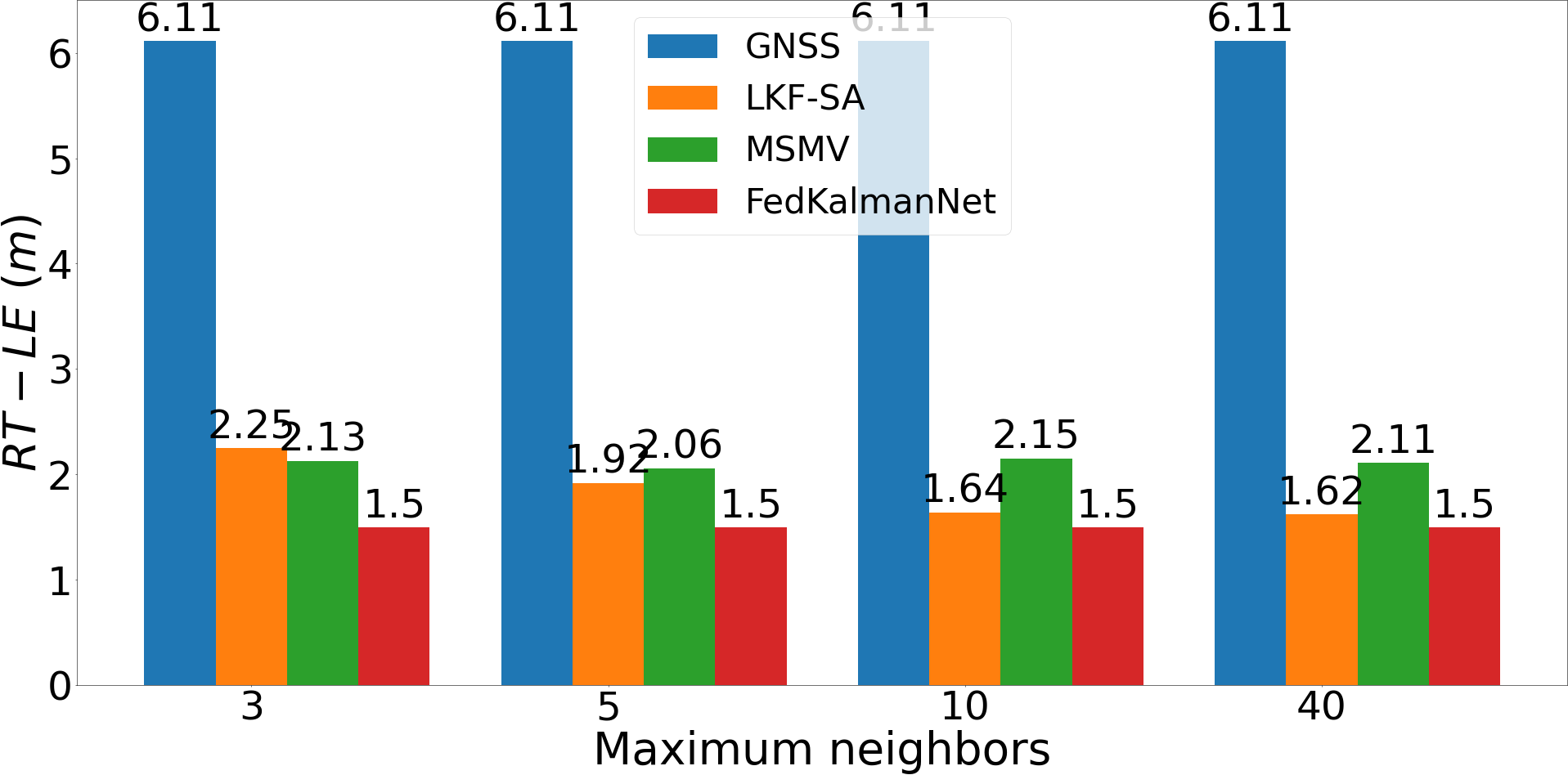}
  \caption{\textbf{FedKalmanNet} outperforms the baseline methods, exploiting only self GNSS and velocity. \textbf{LKF-SA} has to integrate greater amount of information from neighbors to reach \textbf{FedKalmanNet's} accuracy}
  \label{fig:neighbors}
\end{figure}

In the second testing scenario, we will investigate the performance of \textbf{FedKalmanNet} versus CL techniques \textbf{LKF-SA} and \textbf{MSMV}. Our goal is to demonstrate that federated data driven localization which exploits only self measurements of ego vehicle, is capable of outperforming collaborative decision-making solutions which require to fuse information coming from nearby vehicles. As such, by estimating the underlying uncertainty through the proposed collaborative training scheme, we will accurately and cost-efficiently localize ego vehicle. Results are summarized in Fig.~\ref{fig:neighbors}. To simulate relative and self measurements that have to be fused by the ego vehicle, we set standard deviation of distance, azimuth and inclination angle measurement noise equal to $\sigma_d = \ 1 \ m$, $\sigma_{az} = \ \sigma_{in} = \ 4^\circ$, respectively, as well as $\boldsymbol{\Sigma_p} = diag(3.5, 3.5, 3.5)$. Furthermore, each vehicle establishes a connected neighborhood with its nearby vehicles within a range of $30m$, consisting of a maximum number of vehicles (based on shortest distance). In Fig.~\ref{fig:neighbors}, we demonstrate $RT-LE$ of each technique with respect to maximum number of connected neighbors. Clearly, GNSS experiences the lowest accuracy, while both \textbf{FedKalmanNet} and GNSS are actually independent of the number of neighbors. \textbf{LKF-SA} improves its accuracy as the size of neighborhood grows larger, reaching $RT-LE$ equal to $1.62m$, requiring at the same time richer V2X communication resources in order to perform the fusion. \textbf{MSMV} performs more or less the same in all cases ($~2.1m$ of $RT-LE$). We see in all cases that \textbf{FedKalmanNet} significantly outperforms the other solutions, reaching $1.5m$ of accuracy, exploiting only the self measurements (GNSS and velocity) of ego vehicle. On the other hand, \textbf{LKF-SA} has to exploit larger neighborhoods in order to enhance the localization performance.

\section{Conclusion}
\label{conclusion}
This paper has introduced \textbf{FedKalmanNet}, the FL counterpart of \textbf{KalmanNet}, in order to enable a collaborative training paradigm among a group of vehicles, aiming to enhcance vehicle localization through self measurments' uncertainty estimation. The distributed learning scenario among a group of vehicles has been formulated through the ATC strategy, where each vehicle exploits initially its local private dataset to train a local \textbf{KalmanNet}, which is then updated by a global aggregation operation at the server side. Evaluation results in CARLA simulator demonstrate that the FL model features almost similar performance with its centralized counterpart, while significantly outperforms the model trained with the data coming from an individual vehicle. Most importantly, we have shown that the proposed FL data-driven localization framework exploiting only self measurements, performs much more efficient than collaborative decision-making schemes, which fuse data from large neighborhoods of connected vehicles in order to localize ego vehicle. As a future study, we will investigate how different aggregation rules at the server side influence the FL model accuracy.

\bibliographystyle{IEEEtran}
\bibliography{mmsp}

\end{document}